\documentclass[10pt,conference,a4paper]{IEEEtran}
\usepackage{booktabs} % nicely formatted tables
\usepackage{url}
\usepackage{soul,color,graphicx,float,epstopdf}
\usepackage{tablefootnote,textcomp,gensymb}
\usepackage{eurosym,booktabs,multirow}
\usepackage[caption=false]{subfig}
\usepackage{amsfonts} 
\usepackage{amsmath}
\usepackage{algorithmic}
\usepackage{array}
\usepackage{stfloats}
\usepackage{booktabs}
\usepackage{url}
\usepackage{float}
\usepackage{mathtools}
\usepackage{graphicx}
\usepackage{pdfpages}
\usepackage{subfig} 
\usepackage{balance}

\definecolor{Orange}{rgb}{1,0.5,0}

\begin{document}

% \null%
% \includepdf[pages={1,2}]{ICPR 2020 - Response Letter5.pdf}
% \newpage
\title{Deep Transfer Learning for WiFi Localization}
\author{\IEEEauthorblockN{Peizheng Li\IEEEauthorrefmark{1}\IEEEauthorrefmark{2}\IEEEauthorrefmark{3},
Han Cui\IEEEauthorrefmark{3},
Aftab Khan\IEEEauthorrefmark{2},
Usman Raza\IEEEauthorrefmark{2},
Robert Piechocki\IEEEauthorrefmark{3},
Angela Doufexi\IEEEauthorrefmark{3},
Tim Farnham\IEEEauthorrefmark{2},
}\\ 
\IEEEauthorblockA{\IEEEauthorrefmark{2}Bristol Research \& Innovation Laboratory, Toshiba Europe Ltd., UK\\ 
\IEEEauthorrefmark{3}University of Bristol, UK\\
Email: {\{Peizheng.Li, Aftab.Khan, Usman.Raza, Tim.Farnham\}@toshiba-bril.com;}\\
{\{Han.Cui, R.J.Piechocki, A.Doufexi\}@bristol.ac.uk}}}
\maketitle
\begin{abstract}
This paper studies a WiFi indoor localisation technique based on using a deep learning model and its transfer strategies. We take CSI packets collected via the WiFi standard channel sounding as the training dataset and verify the CNN model on  the subsets collected in three experimental environments. 
We achieve a localisation accuracy of 46.55 cm in an ideal $(6.5m \times 2.5m)$ office with no obstacles, 58.30 cm in an office with obstacles, and 102.8 cm in a sports hall $(40 \times 35m)$. Then, we evaluate the transfer ability of the proposed model to different environments. The experimental results show that, for a trained localisation model, feature extraction layers can be directly transferred to other models and only the fully connected layers need to be retrained to achieve the same baseline accuracy with non-transferred base models.
This can save $60\%$ of the training parameters and reduce the training time by more than half.
% Finally, an ablation study of the training dataset shows that, in the office scenario, after freezing the feature extraction layers of the base model, only $70\%$ of the training data are required to obtain an accurate localisation result; on the sports hall dataset, all the training sets are still needed for the re-training.
Finally, an ablation study of the training dataset shows that, in both office and sport hall scenarios, after reusing the feature extraction layers of the base model, only $55\%$ of the training data is required to obtain the models' accuracy similar to the base models.
\end{abstract}

\begin{IEEEkeywords}
WiFi, Indoor localisation, Deep CNN, Transfer learning.
\end{IEEEkeywords}

\section{Introduction}
\label{sec:introduction}
Indoor wireless localisation is the cornerstone of many interactive services, such as remote healthcare, access control, smart homes, asset tracking, and indoor navigation. However, due to the multipath effect in indoor environments, achieving an accurate localisation has long been a problem. Although Ultra-wide band (UWB) and mmWave, due to their larger frequency bandwidths, can partly mitigate the impact of multipath and achieve a relatively high accuracy, the problems of the cost, energy consumption, and the spectrum regulation of these devices hinder a large-scale global adoption. 
On the contrary, WiFi, due to its widespread adoption, low cost, and the standardized protocols, is a good candidate for pervasive indoor localisation. 
In the past two decades, WiFi localisation has undergone a profound evolution. Early localisation technologies include geometric localisation based on RSSI \cite{bahl2000radar}, AoA \cite{farnham2019indoor}, or fingerprinting methods \cite{so2013improved}. These methods suffer from low accuracy or time-intensive data collection process. 
More importantly, inherent time-varying nature of WiFi signals even in a stationary state impacts the performance of these traditional geometric localisation methods \cite{yang2013rssi}. 
In recent years, the progressive exploitation of the channel state information (CSI) enables more fine-grained environmental characterisation through WiFi channel sounding \cite{halperin2010predictable}. The orthogonal frequency division multiplexing (OFDM) scheme adopted in 802.11 provides individual sub-carrier responses for the channel at its carrier frequency, which includes more dynamic features that can be used for localisation. In order to model the relationship between the dynamic changing signal characteristics and the location of the target, machine learning (ML) has recently gained researcher interest. 

ML consists of four main categories: supervised learning, semi-supervised learning, unsupervised learning, and reinforcement learning \cite{goodfellow2016deep}. The main category used in wireless localisation is supervised learning. Although semi or unsupervised learning requires less or no labeled training data, it is difficult to achieve as high accuracy as supervised approaches.
For reinforcement learning, the interaction system between the agent (target to be located) and the environment (radio transmission) is very difficult to model and implement. Therefore, these three schemes are still in an early research stages \cite{li2019deep}.
However, in contrast to the supervised learning used in image, audio or video processing, its application in wireless localisation has two specific problems. Firstly, the data collection, labelling, and pre-processing involves a radio survey of the deployment site. Accurately collecting the ground truth location of the target and correlating with the radio signal features is a hard or labour intensive problem. The second is the transferability of the model, as a model trained in one indoor environment often performs badly when applied to others. Different indoor environments will experience different multipath effect that changes the radio signal characteristics depending on the specific details of the layout and surface properties. Thus, model parameters trained by these characteristics are not generalizable and can trigger incorrect estimation. 
Generally, in a new indoor environment, new training data should be collected to retrain any ML model. These two factors restrict the popularity of supervised learning in wireless localisation.
In order to reduce the resources for retraining the model, we set our sights on transfer learning. That is, transferring some knowledge from a trained localisation model to improve the training efficiency of a new model with a simplified data collection process.

In our previous work \cite{li2020wireless}, we studied the application and performance of a shallow neural network (SNN), a convolutional neural network (CNN) and a long short-term memory (LSTM) model in WiFi localisation. Among these approaches, with a reasonably constructed input, the CNN can directly use the raw CSI data to estimate the target location. The results showed a better performance of CNN than the other two models when considering the training time, generalisability, and accuracy. 
In this paper, we collected more WiFi datasets in different indoor environments, revised the localisation performance of the previously proposed CNN model and studied the transferability of the model.
The main contributions of this paper are summarised below:
\begin{itemize}
\item For the first time, we collected data in both large and small indoor environments to verify the localisation performance of a deep CNN model in different buildings.
\item This paper verifies the transfer learning ability of the model on the different training sets.
\item Furthermore, the key layers of the model, that can be reused by transfer learning, are identified and the time and data volume required for retraining the other layers calculated.
\end{itemize}
% \begin{figure*}[t]   
%     \subfloat[\label{fig:office layout}]{
%       \begin{minipage}[t]{0.315\linewidth} %   \centering   
%         \includegraphics[width=2.25in]{office layout.pdf}   
%       \end{minipage}%
%       }
%       \hfill
%         \subfloat[\label{fig:camera layout}]{
%       \begin{minipage}[t]{0.315\linewidth}   
%         \centering   
%         \includegraphics[width=2.25in]{camera_layout.pdf}   
%       \end{minipage} 
%       }
%      \hfill
%      \subfloat[\label{fig:dingle layou}]{
%       \begin{minipage}[t]{0.315\linewidth}   
%         \centering   
%         \includegraphics[width=2.25in]{dingle layout_v2.pdf}   
%           \end{minipage}  
%       }
%       \caption{(a) The experimental setup of the office environment; (b) Positions of of AP and camera along with target tag moving trajectory; (c) The sport hall layout.} \label{fig:experiment layout}
%     \end{figure*} 

\begin{figure*}[t]   
    \subfloat[\label{fig:office layout}]{
      \begin{minipage}[t]{0.46\linewidth} %   \centering   
        \includegraphics[width=3.2in]{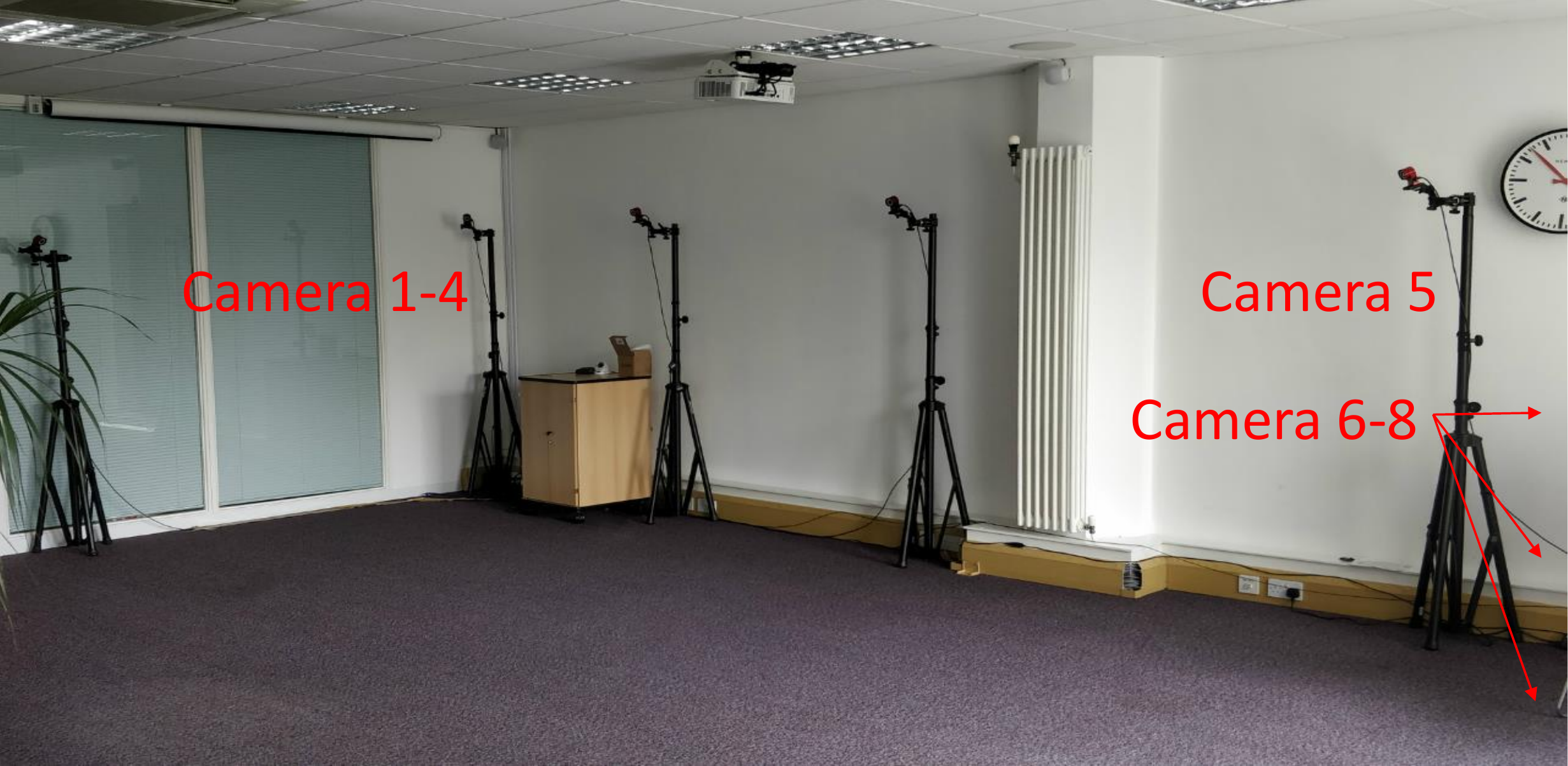}   
      \end{minipage}%
      }
      \hfill
     \subfloat[\label{fig:dingle layou}]{
      \begin{minipage}[t]{0.46\linewidth}   
        \centering   
        \includegraphics[width=3.2in]{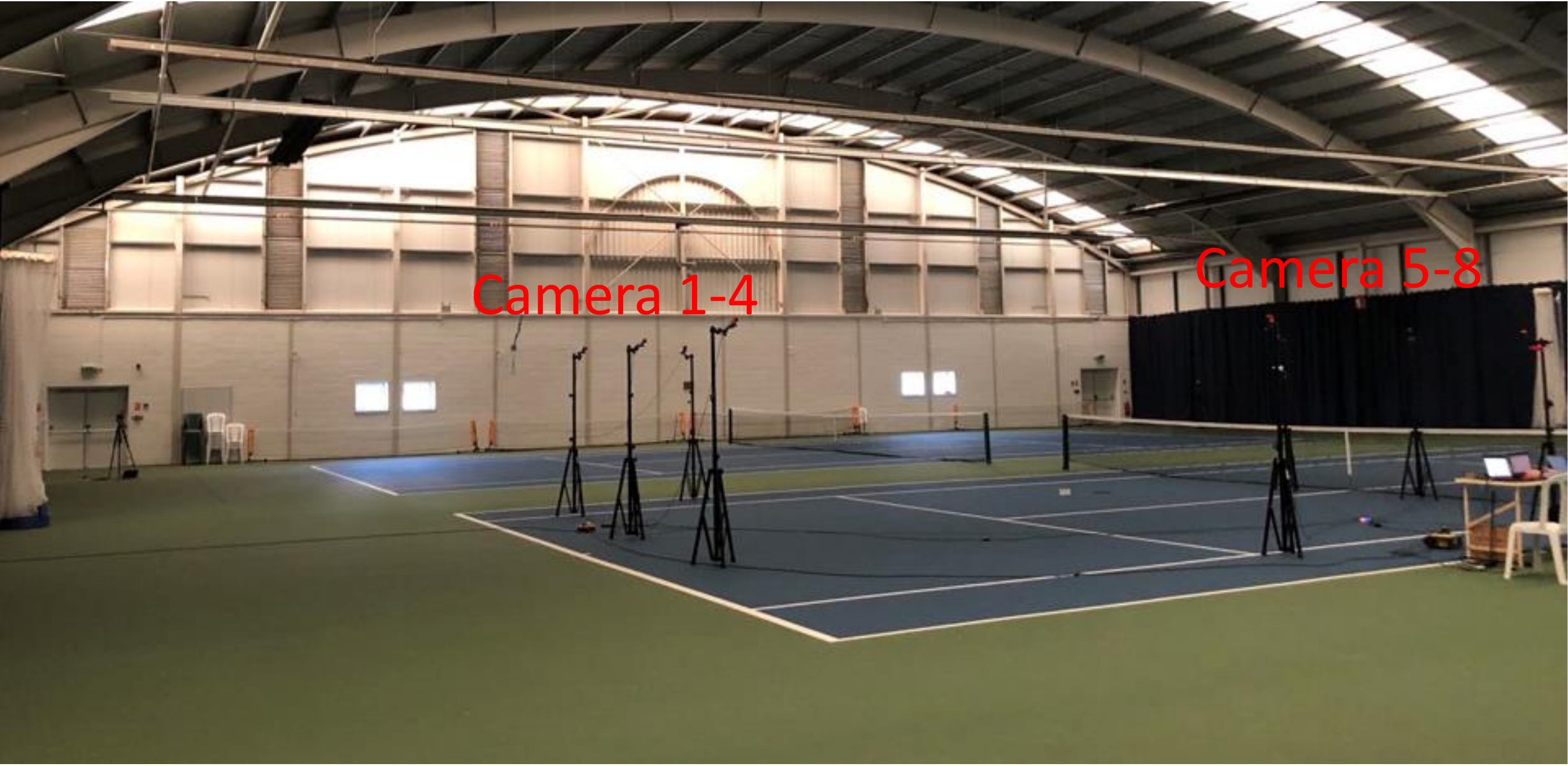}   
          \end{minipage}  
      }
      \caption{(a) The experimental setup of the office environment; (b) The sport hall layout.} \label{fig:experiment layout}
    \end{figure*} 
    
    \begin{figure}[t]
    \centering
    \includegraphics[width=0.43\textwidth]{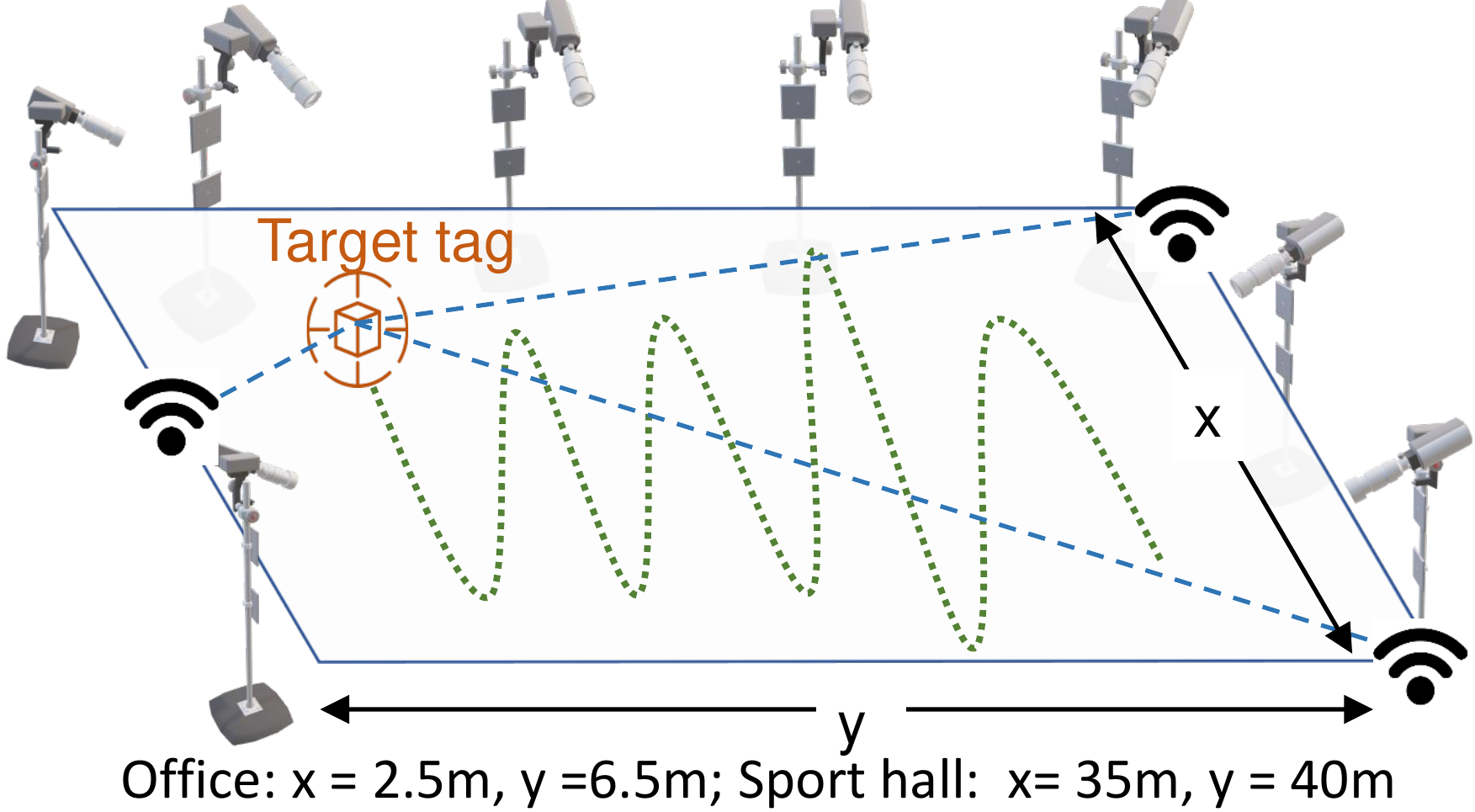}
    \caption{Positions of of APs and cameras along with target tag moving trajectory}
    %\vspace{-0.5cm}
    \label{fig:camera layout}
\end{figure}

    % \subfloat[\label{fig:camera layout}]{
    %   \begin{minipage}[t]{0.315\linewidth}   
    %     \centering   
    %     \includegraphics[width=2.25in]{camera_layout.pdf}   
    %   \end{minipage} 
    %   }
    %  \hfill
\section{Background}
\label{sec:Background}
With the emergence of CSI extraction tools in recent years, the characteristics used for WiFi localisation have gradually shifted from the received signal strength indicator (RSSI) to CSI.  The RSSI only provides the combined signal strength subject to multipath superposition. In contrast, the CSI is in the form of quantified channel frequency response (CFR), which captures the propagation characteristics of different subcarriers from transmitters to receivers \cite{ma2019wifi}.  The CSI provides more rich information that can be used for localisation. With the NIC Intel 5300 card used in this paper, each received CSI packet can report the CFR of 30 subcarriers. This can be represented as:
\begin{equation}\label{formula: H}
{H}_{i}=|{H}_{i}|{{e}^{j\sin \{\angle {{H}_{i}}\}}, i\in \left [ 1,30 \right ] }
\end{equation}
where $|{H}_{i}|$ and $\angle {{H}_{i}}$ are the amplitude and the phase response of the subcarrier $i$ respectively. 
These measurements of amplitudes and phases are directly influenced by the environment and thus capture its characteristics to a certain extent. But, it is not obvious to deduce the target position because of the additional  amplitude and phase distortion caused by hardware imperfections and multipath effect.
To reduce interference, Kotaru et al. \cite{kotaru2015spotfi} adopt an AoA based geometric method. DeepFi \cite{wang2015deepfi} and PhaseFi \cite{wang2015phasefi} construct neural networks by using amplitude and phase respectively. Wang et al.\cite{wang2018deep} utilise AoA for constructing images. These methods still require manual CSI feature extraction. In contrast, we deploy in our earlier work \cite{li2020wireless} a deep CNN in which each input data sample is constructed from a combination of CSI packets after amplitude and phase processing.

In addition, transfer learning is a useful tool to solve the problem of insufficient training datasets in machine learning. It can transfer the knowledge from a different but relevant domains to a target domain to reduce the dependence of the retraining on the large amount of data. Transfer learning has been successfully applied in natural language processing, especially in the application of multilingual translation \cite{johnson-etal-2017-googles}.
Tan et al. \cite{tan2018survey} divides transfer learning into four categories: instance-based, mapping-based, network-based, and adversarial-based.
The wireless localisation studied in this paper exploits deep CNN to explore the spatial characteristics of CSI in an environment. An unavoidable problem of this approach, in the deployment stage, is that different indoor environments will have different spatial characteristics. Hence, the generalizability of a model trained in a particular environment will become very poor. Thus whole model or a part of the layers is to be retrained. This paper studies the network-based transfer learning techniques, in the target domain, with different indoor environment characteristics. We reuse a part of a pre-trained network from the source domain, and assess its feasibility and performance in a target domain. 
The research of the transferability was carried in two aspects with either minor or major changes in the environments. 
We analyse each layer and assess which layers from the source model can be reused without retraining, and evaluate the change in the training time and the training data size requirement. This provides guiding principles for the promotion and model reuse of WiFi localisation based on deep learning in different environments.
% Section \ref{sec:exp} gives the details of the corresponding experiment design and data collection.
 \begin{figure*}[t]
    \centering
    \includegraphics[width=0.80\textwidth]{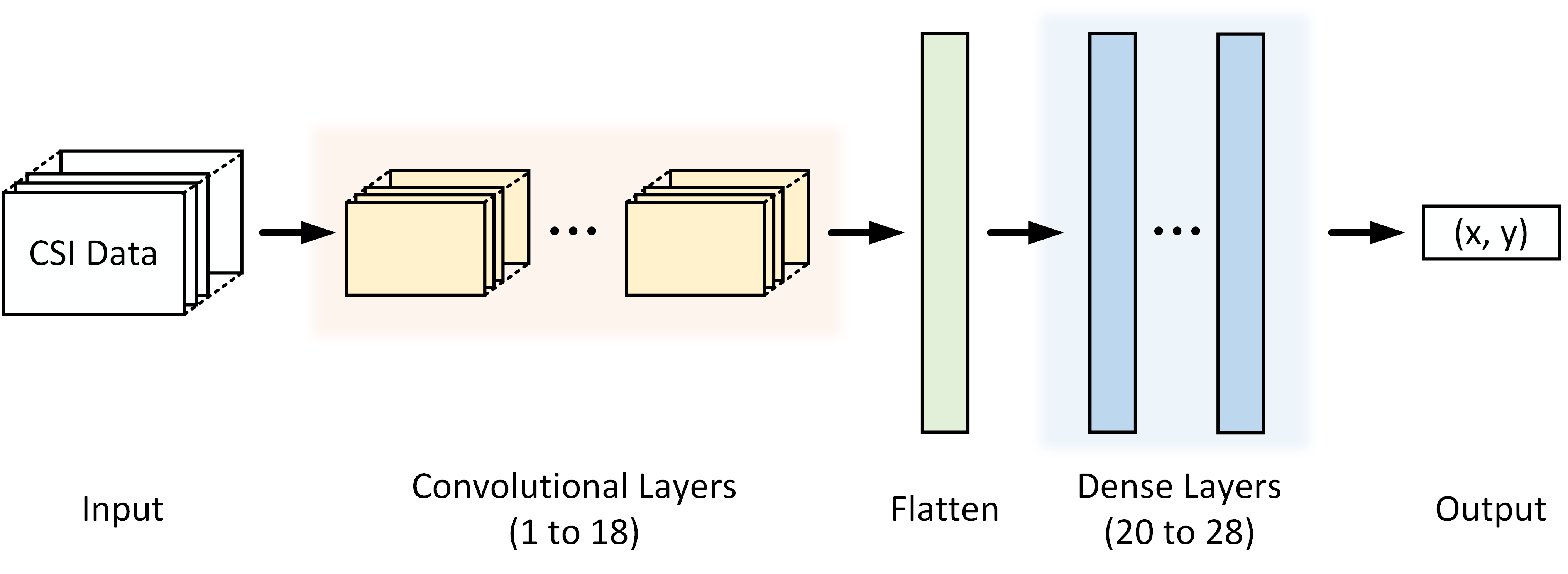}
    \caption{The CNN structure and layer definitions, where the input is a combination of CSI packets, and the output is the the (x,y) coordinates of the target location.}
    %\vspace{-0.5cm}
    \label{fig:CNN_structure}
\end{figure*}
\section{Experimental Setup}
\label{sec:exp}

\subsection{Environment setup and Data Collection}
The training data is collected in two different environments, a $6.5 m \times 2.5 m$ office as shown in Figure \ref{fig:office layout}, and a large $35 m \times 40 m$ indoor sports hall as shown in Figure \ref{fig:dingle layou}. In each room, three commodity WiFi APs are placed in the experimental area (their relative positions shown in Figure \ref{fig:camera layout}). In the data collection period, these APs simultaneously receive broadcast channel sounding packets from the target transmitter, while the target tag making curvilinear motions in the test area. At the same time, the ground truth trajectory is recorded by an OptiTrack system \cite{optiTrack} in the form of Cartesian coordinates at $120$ Hz.
In the office environment, the OptiTrack system is able to cover the whole experimental test area.
In the sport hall, in order to ensure the effectiveness of the groundtruth data collection, we divided this site into several blocks and collected data in different blocks separately. The target object was carried by a person, who continuously moved around in the test area to mimic a severe shadowing environment. The channel sounding packet rate is set to $500$ Hz and the CSI is collected over $30$ subcarriers, with one transmitting antenna and three receiving antennas at each AP. We used conventional antennas in the office and high gain omni-directional antennas in the sports hall to ensure the integrity of the sounding packets. 
\subsection{Dataset}  
\label{sub:dataset}
The dataset used in this paper consists of three sub-datasets. Each sub-dataset contains five independent sessions, and each session's data lasts two minutes. The sub-dataset 1 and 2 (referred as office 1 and office 2) are collected in the same office but with different anchor AP positions, and with more shading objects in the second one. The sub-dataset 3 is collected in the middle area of the sport hall. Every sub-dataset contains around 12 thousand
labelled samples.
\subsection{Transfer Learning in CSI Data}
\label{sec:method}
Our proposed CNN, as shown in Figure \ref{fig:CNN_structure}, is mainly composed of three parts: 18 convolutional layers (layer 1 to 18), a flatten layer (layer 19) and 9 fully connected layers (layer 20 to 28). Between the convolutional layers 1-18, a pooling layer is inserted, that is layer 15. The convolutional layers extract the features of the input data; the pooling layer reduces the feature dimensions and removes non-significant information; the flatten layer converts multi-dimensional input into one dimension output and feed it to fully connected layers; the fully connected layers transform the extracted features to the label space.
For the sounding packets, captured at the three APs, we construct a specific data input pattern in which all the collected amplitudes and phases of the subcarriers are integrated into a $75*30*6$ tuple, which encodes the characterized channel state of the target location and is fed as the input to the CNN model. The details of each layer can be seen in \cite{li2020wireless}.

Even if the sounding packets are collected from different rooms with different sizes, layouts and features, the data is expected to still have some common hidden features. For example, the amplitude and phase of adjacent subcarriers should maintain a certain degree of continuity and stability, and these values should be limited within a certain range. 
The convolutional layers of the network are expected to be able to extract these hidden features from the initial tuple and pass them to the subsequent layers. 
This assumption forms the basis for transfer learning to be applied.
The feature extraction ability of the trained convolutional layers can be transferred to a new network and reused, to reduce the training consumption of the new network. 
The indoor multipath effect can be overcome by tuning the key layers other than the feature extraction layers.
In the next section, we evaluate which layers are critical for common features extraction, which layers need fine-tuning for an accurate localisation, and how much data tuning is needed. 
For a trained base model, when applying to a new data set from another environment, we freeze the base model layer by layer to be reused in the new model, then train the remaining unfrozen layers with the new dataset to evaluate the importance of each layer.
\section{Performance Evaluation}
\label{sec:Results}

% Using sub-dataset office 1, we trained this CNN model and treat it as the base model, where the total trainable parameters of this model is $6,265,180$, and training process costs about 40 minutes. 

% The new data set comes from two places: (1) In the same office, but new occlusions are added and the anchor APs positions are changed. (2) A sports hall. All data are collected in the manner described in section \ref{sec:exp}. 
\begin{figure*}[t]   
    \subfloat[\label{fig:transfer1}]{
      \begin{minipage}[t]{0.45\linewidth} %  \centering   
        \includegraphics[width=3.2in]{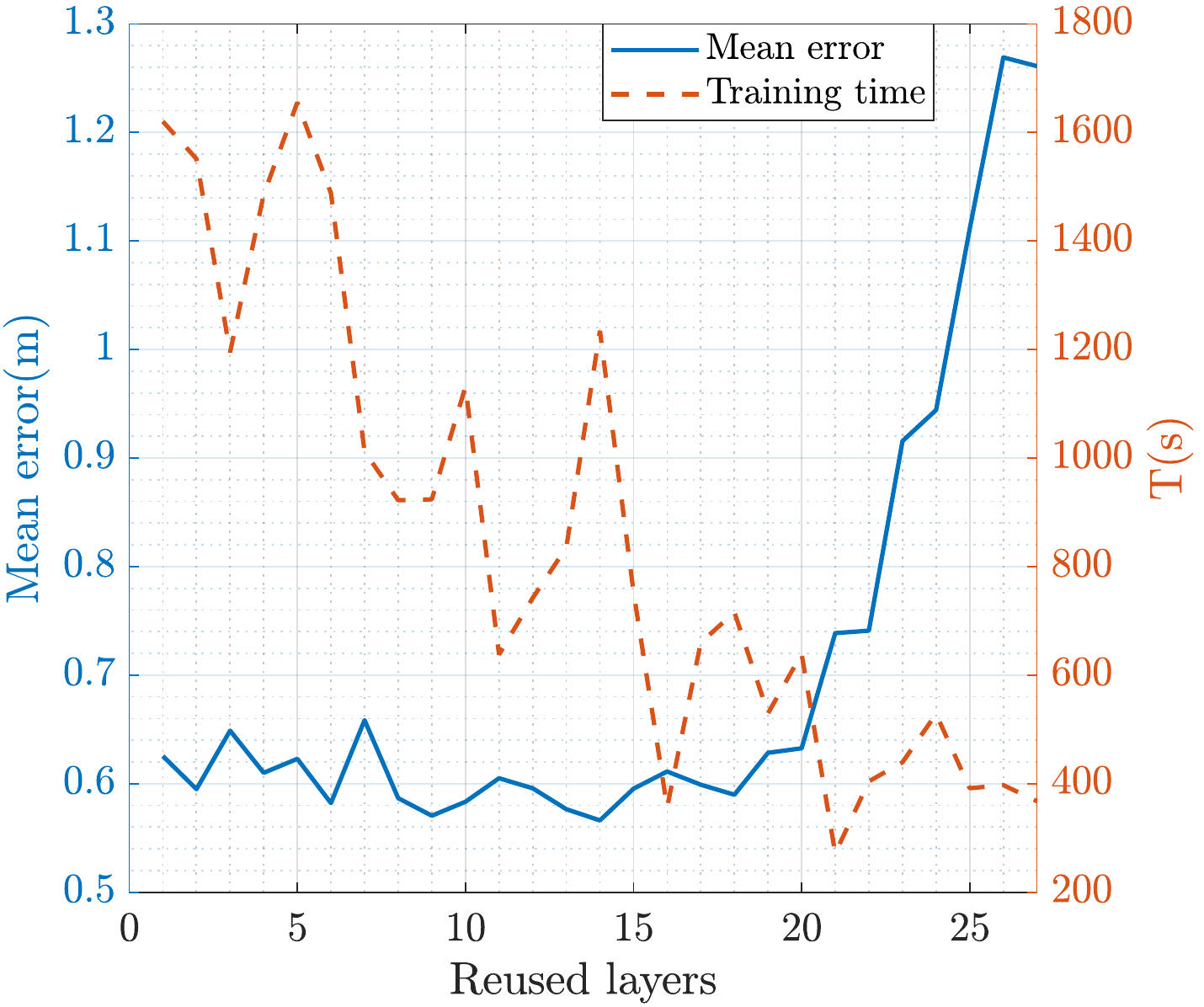}   
      \end{minipage}%
      }
      \hfill
        \subfloat[\label{fig:transfer2}]{
      \begin{minipage}[t]{0.45\linewidth}   
        \centering   
        \includegraphics[width=3.2in]{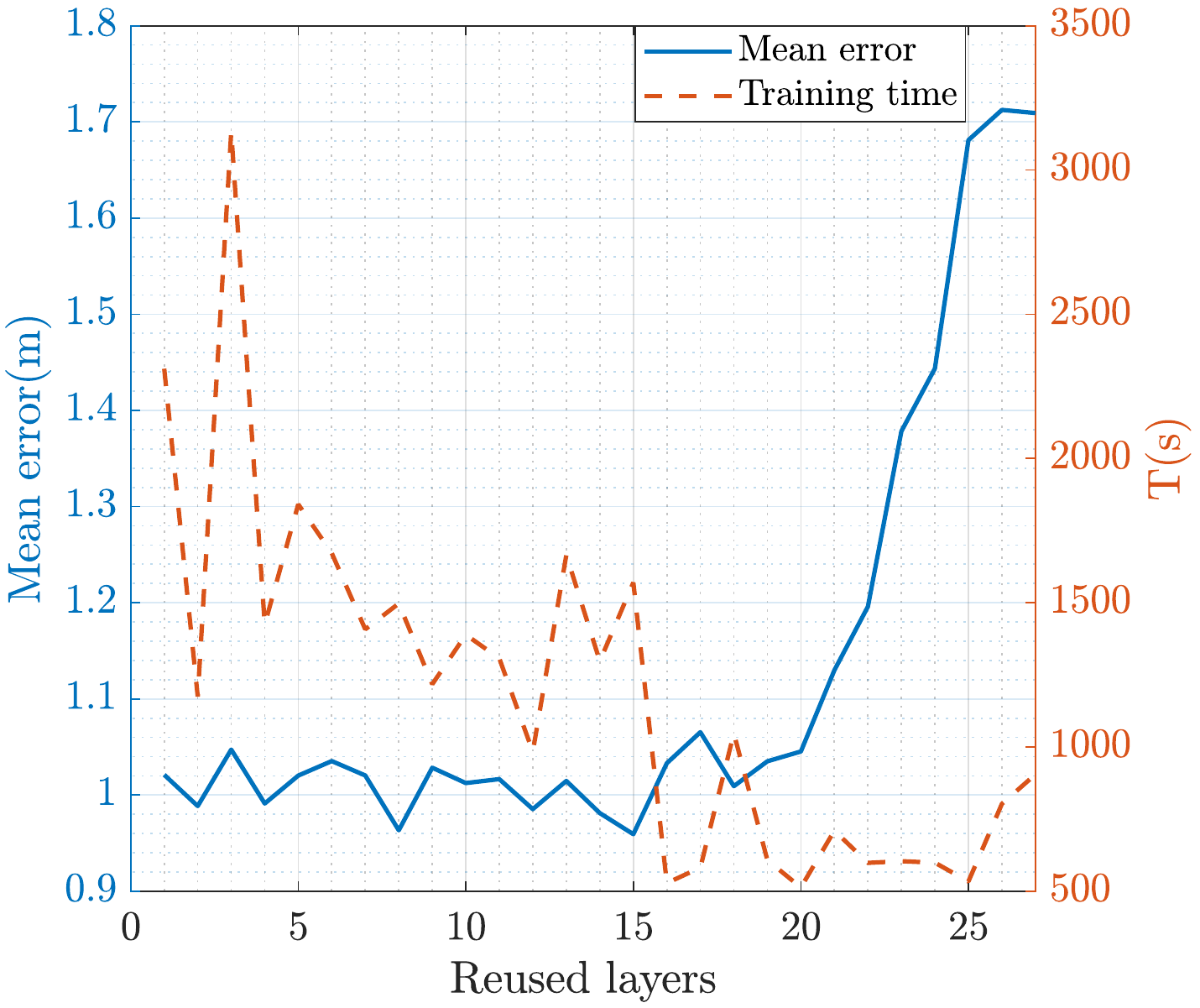}   
      \end{minipage} 
      }
      \caption{The performance of the transferred model from office 1 to office 2 (fig a) and the sports hall (fig b).} \label{fig:transfer}
    \end{figure*} 
The experimental computer used in this paper has an i7-9700F CPU @ $3.0$ GHz, $16$ GB of memory and a Nvidia RTX2070 Super GPU with $8$ GB of dedicated memory. The model training is performed on Keras using TensorFlow 2.1 as the backend. The loss function adopted is the mean absolute error (equation \ref{eq:mae}).
\begin{equation}
loss=mean(|{{(x,y)}_{true}}-{{(x,y)}_{pred}}|)
\label{eq:mae}
\end{equation}
We use the Adamax optimizer \cite{kingma2014adam} with a batch size of $30$ and an adaptive learning rate. The selu activation function is applied to all intermediate layers. In order to evaluate the performance of the model, we use the Euclidean distance $e$ between the estimated location $(x,y)$ and the ground truth location ${(x,y)}_{true}$ (equation \ref{eq:indicator}). 
\begin{equation}
e = \sqrt{({x}_{true}-x)^2+({y}_{true}-y)^2}
\label{eq:indicator}
\end{equation}
\subsection{Model Performances on different datasets}
We used the three data subsets collected from different indoor environments as described in Section \ref{sub:dataset}.
Each data subset contains $5$ independent sessions. We used $3$ sessions as training sets, the other two as test sets and validation sets respectively, and finally the localisation accuracy of validation sets was used to evaluate the model.
We trained three independent CNN models and applied a 5-fold cross-validation on each model, and the early stopping was used to halt the training which was triggered after $10$ no improvement training epochs. The localisation mean error on each sub-dataset is given in table \ref{table:performance}.
\begin{table}[t]
\centering
\caption{WiFi localisation mean error in different indoor environments.}
\label{table:performance}
\begin{tabular}{cccc}
\toprule
Scenarios & Office 1 & Office 2    & Sport hall   \\ \midrule
Mean error (m) & 0.4655     & 0.5830 & 1.0279 \\ \bottomrule
\end{tabular}
\end{table}
It can be seen that the base model in office 1 has the best accuracy of $0.4655m$.
The performance in office 2 is reduced due to the stronger shadowing. In the sports hall, because the experimental site is much larger than the office, the amplitude and phase characteristics of the WiFi subcarriers change with respect to the distance become less significant, so the CNN extracts fewer features, and the localisation accuracy of the model becomes worse. We set these values as the reference to discuss the transferability of the model. 
\subsection{Transfer ability}
\label{subsec: transfer ability}

% even in which there are $7$ layers, such as pooling layers, dropout layers do not contain trainable parameters, we still take these layers in the analysis.

The model has a total of $28$ layers and about $6$ million parameters. We used the model trained in office 1 as the base model and reused a part of its parameters during retraining.
The number of reused, or frozen, layers was gradually increased from only the first layer to all layers except for the $28 th$ layer. 
During this process, as we fed the data of the data subset 2 and 3 to retrain the model, the localisation accuracy and the training time of the model are shown in Figure \ref{fig:transfer}.
Although, due to the inconsistent direction of the gradient descent every time the model is retrained, the time required for the new model to retrain when some layers are reused has no accurate quantitative meaning (see the red dotted line). However, it can still be seen qualitatively that the turning point of the model training time appears to be layer 15, the pooling layer, while the turning point of the model accuracy lies on layer 19, the flatten layer. 
This observation indicates that, for a well-trained localisation model, its layers before the flatten layer own certain common knowledge about the data and can be transferred to other models without retraining or tuning, whereas the fully connected layers' parameters cannot be utilised directly. 
In other words, the convolutional layers and the pooling layers play the role of feature extraction, while the fully connected layers further fit the data and perform the prediction. 
On the other hand, it can be seen that the majority of the training time consumption lies on the convolutional layers. 
The training of the fully connected layers only involve %$203,335,4$ {\color{red}(?)} 
around $2$ million parameters and can save about half of the training time.
It is noticeable that, as the red lines indicate, the training time of the data subset 3 is longer than that of data subset 2, which shows that the WiFi data collected in a larger environment has weaker spatial correlation characteristics and requires more training epochs.
\subsection{Data ablation}
\label{subsec: Datasize ablation}
\begin{figure}[t]
    \centering
    \includegraphics[width=0.43\textwidth]{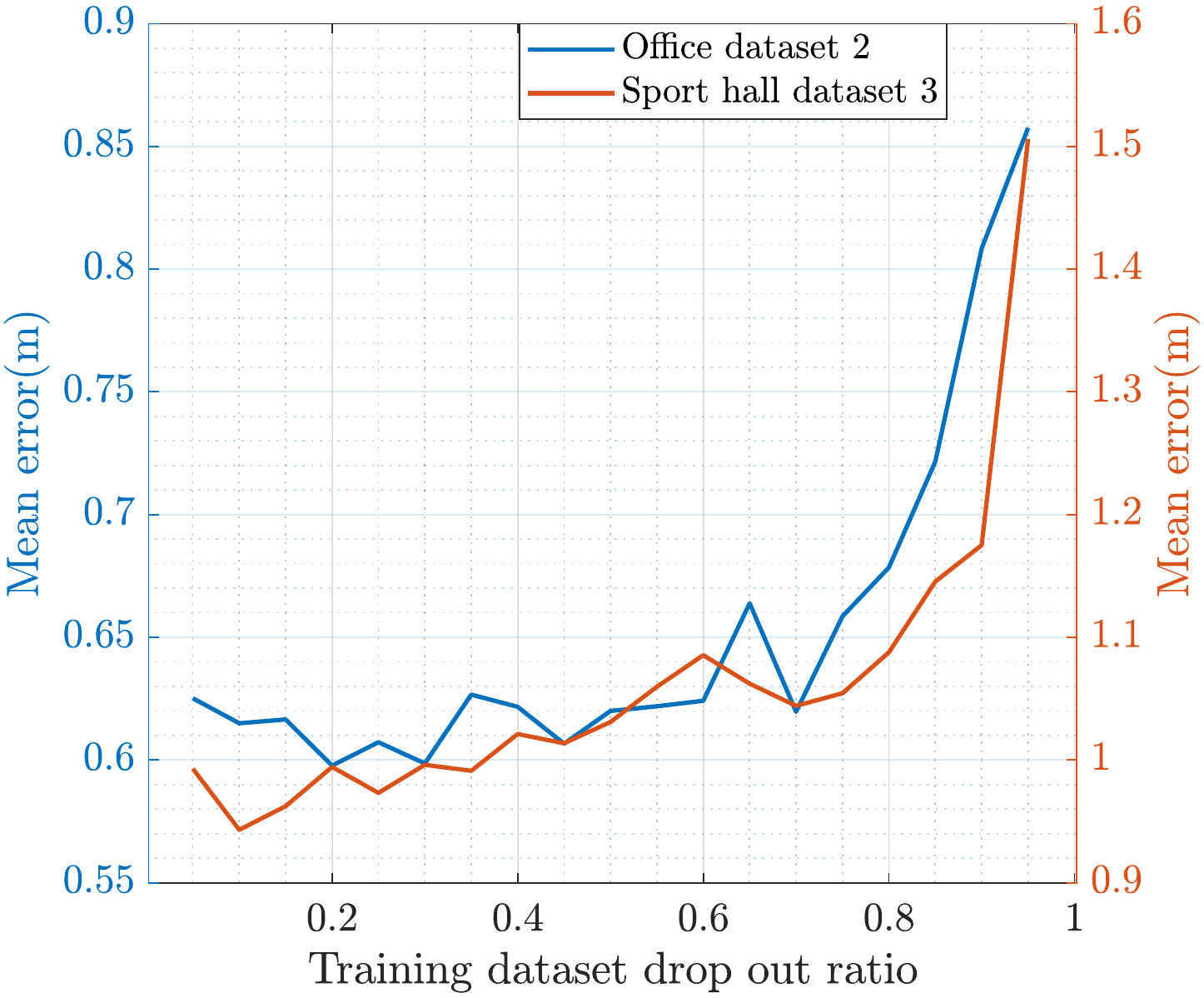}
    \caption{The final localisation accuracy of the two models, originating from the base model with all convolutional layers frozen and the remaining layers re-trained with reduced training data.}
    %\vspace{-0.5cm}
    \label{fig:ablation}
\end{figure}

In this section we study the importance of the data size to the model retraining. 
We freeze the first 19 layers of the base model (all layers before the error escalates) and evaluate the impact of the training data size on the final localisation accuracy for the office 2 and the sports hall dataset. 
We randomly dropped a proportion of data in the dataset and trained the model with the rest of the data.
The results are shown in Figure \ref{fig:ablation}, in which the data drop out range is set to $[0.05, 0.95]$ with an interval of $0.05$. 
It can be seen that dropping out 45\% of the training data has little effect on the accuracy of the new models. Therefore, it can be considered that with the help of transfer learning, by continuing to use the convolutional layer parameters of the base model, the training of the new models can save about 45\% of the training data, no matter in the office or the sport hall.

% When only the fully connected layers were to be retrained, % about 30\% of the training data in office sub-dataset 2 do not contribute to the training, whereas for the sports hall almost all the training data contributed {\color{red}(where this conclusion comes from?)}. 
% On the other hand, with the help of transfer learning, only 40\% of the data can train the model to achieve relatively acceptable localisation accuracy {\color{red}(again?)}.

\section{Conclusions}
\label{sec:Conclusion}
This paper studies the transfer learning strategy of a deep CNN based WiFi indoor localisation technique. 
In addition to our previous work, we collected two WiFi CSI datasets in real-world scenarios and verified the performance of our CNN model, with an accuracy of $0.58$ m and $1.03$ m in a regular office and a sports hall respectively. 
We discussed the transfer learning application between different environments. 
The results show that the convolutional layers from a trained CNN base model can be directly transferred to a new model without retraining, whereas the fully connected layers need to be retrained.
A transferred model can reduce the model training time by around 50\% and save up to 45\% of the training data.
\bibliographystyle{IEEEtran} %
\balance
\bibliography{IEEEabrv,refs} 
\end{document}